\documentclass[10pt,twocolumn,letterpaper]{article}

\usepackage{cvpr}
\usepackage{times}
\usepackage{epsfig}
\usepackage{graphicx}
\usepackage{amsmath}
\usepackage{amssymb}
\usepackage{shortbold}
\usepackage{booktabs}
\usepackage{enumitem}
\usepackage{color}
\usepackage{caption}
\usepackage{bbm}
\usepackage{setspace}
\usepackage[normalem]{ulem}
\definecolor{DarkGray}{gray}{0.3}

\usepackage[pagebackref=true,breaklinks=true,letterpaper=true,colorlinks,bookmarks=false]{hyperref}

\cvprfinalcopy 

\def\cvprPaperID{506} 

\ifcvprfinal\fi
\begin{document}

\title{Designing Deep Networks for Surface Normal Estimation}

\author{Xiaolong Wang,  David F. Fouhey,  Abhinav Gupta  \\
The Robotics Institute, Carnegie Mellon University\\
{\tt\small \{xiaolonw, dfouhey, abhinavg\}@cs.cmu.edu}
}

\maketitle


\begin{abstract}
In the past few years, convolutional neural nets (CNN) have shown incredible promise for learning visual representations. In this paper, we use CNNs for the task of predicting surface normals from a single image. But what is the right architecture we should use? We propose to build upon the decades of hard work in 3D scene understanding, to design new CNN architecture for the task of surface normal estimation. We show by incorporating several constraints (man-made, manhattan world) and meaningful intermediate representations (room layout, edge labels) in the architecture leads to state of the art performance on surface normal estimation. We also show that our network is quite robust and show state of the art results on other datasets as well without any fine-tuning.
\end{abstract}

\vspace{-0.2in}
\section{Introduction}
\vspace{-0.05in}
The last two years in computer vision have generated a lot of excitement:
deep convolutional neural networks (CNNs) have broken the barriers of performance on
tasks ranging from scene classification to object detection and fine-grained
categorization. For instance, on object detection, performance on the standard
dataset has gone up from a mAP of 33.7 to 58.5 in just two years.
While CNNs have been shown tremendous success on semantic tasks such as
detection and categorization, their performance on other vision tasks
such as 3D scene understanding and establishing correspondence has been not
as extensively studied.


Recently, Eigen et al.~\cite{Eigen14} presented a deep convolutional network
approach to estimating depth from a single image. This model treats the depth
prediction as a regression problem and uses a feed-forward convolutional network for the task.  Specifically, they presented a new two-level architecture
where the coarse level architecture predicted the coarse layout and the finer network used
the output of the coarse network to predict the finer resolution layout. They demonstrate
the effectiveness of deep networks by achieving state of the art performance on the
task of depth prediction.

In this paper, we want to explore the effectiveness of deep networks on the
task of predicting surface normals from a single image. The most straightforward approach
would be use the architecture similar to Eigen  et al.~\cite{Eigen14} but
regress to 3D surface normal space. But such an approach would disregard over
five decades of work in 3D scene understanding from the early blocks world~\cite{Roberts65} and
line-labeling~\cite{Huffman71,Clowes71,Kanade80} work to recent investigations into similar ideas in a data-driven
era\cite{Hedau09,Lee10,Gupta10,Schwing12,Zhao13,Fouhey14c}.  Instead
in this paper, we want to ask a basic question: are there lessons we have
learned from previous research that we can borrow and apply in designing deep
networks for the task of surface normal estimation?

\begin{figure*}[t]
\includegraphics[width=\textwidth]{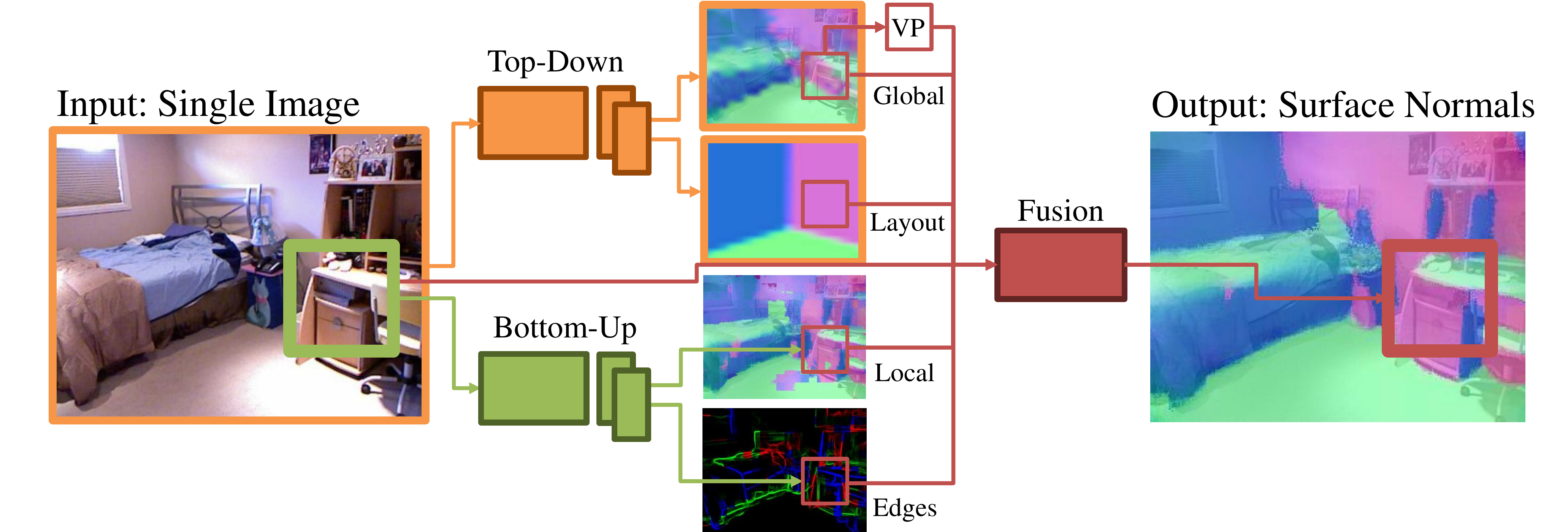}
\caption{{\bf An overview of our approach} to predicting surface normals of a scene from a single image.
We separately learn top-down and bottom-up processes and use a fusion network to fuse the contradictory
beliefs into a final interpretation. {\bf Top-down processes:} our network predicts a coarse $20 \times 20$ structure
and a vanishing-point-aligned box layout from a set of discrete classes. {\bf Bottom-up processes:} our network
predicts a structured local patch from a part of the image and line-labeling classes:
\textcolor{blue}{convex-blue}, \textcolor{green}{concave-green}, and \textcolor{red}{occlusion-red}.  {\bf Fusion process:} our network fuses the outputs of the two
input networks, the rectified coarse normals with vanishing points(VP) and images to produce substantially better results.
}
\vspace{-0.2in}
\label{fig:overview}
\end{figure*}

We propose to inject the hard-won insights about 3D representations and
reasoning into the deep learning framework for surface normal predictions. We
argue that while deep networks have been particularly successful in learning
image representations, their design can benefit greatly by from
past research in 3D scene understanding. We show that incorporating this
knowledge in designing deep networks lead to {\bf state of the art
performance} in surface normal estimation. We additionally  show a 7-8\% improvement over
a standard feed-forward network. More importantly, such networks provide coherent and
deeper understanding in terms of surface normals, room layout and edge labels (Figure~\ref{fig:overview}).


\subsection{Design Decisions and Contributions}
So, what have we learned that could be useful in designing deep networks? Over the past decade, we see three themes on 3D scene understanding emerge again and again:

\noindent {\it \bf Fusing top-down and bottom-up:} A quick
glance at scenes suggests that a single approach is unlikely to work everywhere: the orientation of window blinds, cabinets and tiled floors can be recognized from local cues alone; but blank walls and textureless surfaces require guidance from their context and from top-down models.

Following this, instead of having a single-feed forward network we design a
three model architecture. We build bottom-up and top-down networks, and learn a
fusion network that produces results that are much better than either network
alone (See Figure~\ref{fig:overview}). Our fusion network can be viewed as a
form of learned reasoning that replaces previous optimization-based attempts to
reconcile evidence \cite{Gupta10,Lee10,Schwing13,Fouhey14c} from conflicting
sources.

\noindent {\it \bf Human-centric constraints:}
Another set of strong constraints which have emerged over the past decade are
related to the man-made nature of scenes. For instance, it is common for there
to be three orthogonal directions in the scene, or the Manhattan-world
assumption. This led to a great deal of work focused entirely on the
Manhattan-world: e.g., \cite{Hedau09,Lee09,Lee10,Schwing12,Schwing13,Fouhey14c,JXiao14}. Another
constraint in the similar vein is modeling the room layout as an inside-out
box. Inspired by these observations, we include the room layout constraint
while learning the top-down network and we use both room layout and
vanishing point estimates in the fusion network. Our results show that both
constraints lead to improvement.

\noindent {\it \bf Local structure:} Another theme that has emerged in the
past~\cite{Sugihara86,Huffman71,Clowes71,Kanade80} and recently ~\cite{Hoiem07,Karsch13,Fouhey14c} is the reasoning between
surface normals and the edges in the images. Inspired by these local
constraints, we incorporate them in learning of local network and as an input
in fusion network. We demonstrate inclusion of convex, concave and occlusion
edges improve the performance over the simple feed-forward network.

\vspace{-0.05in}
\section{Related Work}
\vspace{-0.05in}
\label{sec:related}
The topic of 3D  understanding goes back to the beginning of computer vision,
starting from the first thesis, Roberts' Blocks World~\cite{Roberts65}. At the heart of this problem are two related questions:
(1) What are the right primitives for understanding? and (2) Given the local
evidence, how can you obtain global 3D scene understanding?

On the topic of primitives, there has been lot of work in the
past~\cite{Roberts65, Waltz75}. For example, Biederman introduced geons as
primitives for scene understanding~\cite{Biederman87}. These geons are
volumetric primitives such as cuboids, cones, cylinders etc. Recently, lot of
work has focused on using edges~\cite{Lee09}, super pixels~\cite{Saxena05} or
segments~\cite{Hoiem05} as primitives for reasoning. But while these
edge/segment based primitives have shown a lot of promise, they are still
plagued by local ambiguities.

There are two ways to resolve ambiguities: (a) perform global reasoning; (b)
design better primitives. On the first front of reasoning, there has been a
significant amount of
work~\cite{Hedau09,Lee10,Gupta10,Schwing12,Zhao13,Schwing13,Fouhey14c}. Most of
these reasoning approaches are based on higher-order volumetric representations
(e.g., room should be inside out box or two volumes should not intersect with
 each other). The second front has been to design better primitives and improve
estimation of likelihoods. A significant step in this direction was proposed by
Fouhey et al.~\cite{Fouhey13a}. The central argument was that that relying on
human intuitions is not the right way of estimating likelihoods and instead the
data itself should be used for the task. 
%

In this work, we take this a step further. Instead of using primitives on
manually designed features such as HoG~\cite{Dalal05}, we use the data to
derive representation right from the pixel. Inspired by the recent success of
Convolutional Neural Networks~\cite{lecun90} on the task of image
classification~\cite{imagenet12}, object
detection~\cite{Girschik14,overfeat14}, depth estimation~\cite{Eigen14}, pose
estimation~\cite{toshev14} etc., we propose to adapt CNNs to learn
representations and primitives for 3D scene understanding. However, instead of
just using a feed-forward network blindly, we propose to design our network
based on the decades of experience in the field of 3D scene understanding.
Specifically, our approach brings together the two threads (representation and
reasoning) in the deep network framework to solve the task.

\vspace{-0.05in}
\section{Overview}
\vspace{-0.05in}
\label{sec:overview}
This paper aims to combine the knowledge gleaned over the past decade in
single-view 3D prediction with the representation-learning power of
convolutional neural networks. Our overall objective motivation is to frame the
single-view 3D problem so that the structure we know is captured and
convolutional networks can do what they do best -- learn strong mappings from
visual data to labels.

Following the lessons we described in the introduction, we build a network with
the following architecture (illustrated in Figure~\ref{fig:overview}).  We
start with two networks: a top-down network that takes the whole image as input
and predicts a coarse global interpretation (Section \ref{sec:topdown}); and a
bottom-up network that acts on local patches in a sliding-window fashion and
maps them to local orientation (Section \ref{sec:bottomup}). Because the
top-down and bottom-up processes have complementary errors, we combine their
output with a fusion network that learns how to incorporate their predictions
(Section \ref{sec:fusion}). Each input network obtains strong performance by
themselves, but by combining them, we obtain substantially better results, both
quantitatively and qualitatively.

In addition to performing top-down/bottom-up fusion, we inject global human-centric
constraints (including room layout, vanishing point) and local surface/edge
constraints into the framework by introducing additional tasks. Our top-down
network predicts room layout as well, and our bottom-up network predicts an edge
labeling. Integrating these extra tasks leads to a more robust final network.
We evaluate our approach in Section \ref{sec:experiments} and analyze what
aspect of our designs gives what types of performance increases.


\vspace{-0.05in}
\section{Method}
\vspace{-0.05in}
\label{sec:method}
We now describe each of the components of our method. For each,
we describe their inputs, outputs, the intermediate layers, and the loss function they minimize.

\vspace{-0.05in}
\subsection{Output: Regression as Classification}
\vspace{-0.05in}
The outputs for the top-down and bottom-up network are: surface normal for each
pixel, room layout and edge labels. The edge label (convex, concave, occluding,
no-edge) is a discrete output space and can be formulated as a
classification problem. However both surface normal and room layout are continuous
output spaces and therefore need to be formulated as regression problems.
We instead use discrete output spaces and perform classification instead of regression.

\noindent {\bf Surface Normal:} We use the surface normal triangular coding technique from Ladicky
et al. \cite{Ladicky14b} to turn normal regression into a classification
problem. Specifically, we first learn a codebook with k-means and a Delaunay
triangulation cover is constructed over the words. Given this codebook and
triangulation, a normal can be re-written as a weighted combination of the
codewords in whose triangle it lies. At training-time, we learn a softmax
classifier on the codewords. At test-time, we predict a distribution over
codewords; this is turned into a normal by finding the triangle in the
triangulation with maximum total probability, and using the relative
probabilities within that triangle as weights for reconstructing the normal.

\noindent{\bf Room Layout:} Room layout is continuous structured output
space. We reformulate the problem as classification by
learning a codebook over box layouts. The codewords are learned with
k-medoids clustering over 6000 room layouts; each codeword is a category
for classification.

\begin{figure*}
\includegraphics[width=\textwidth]{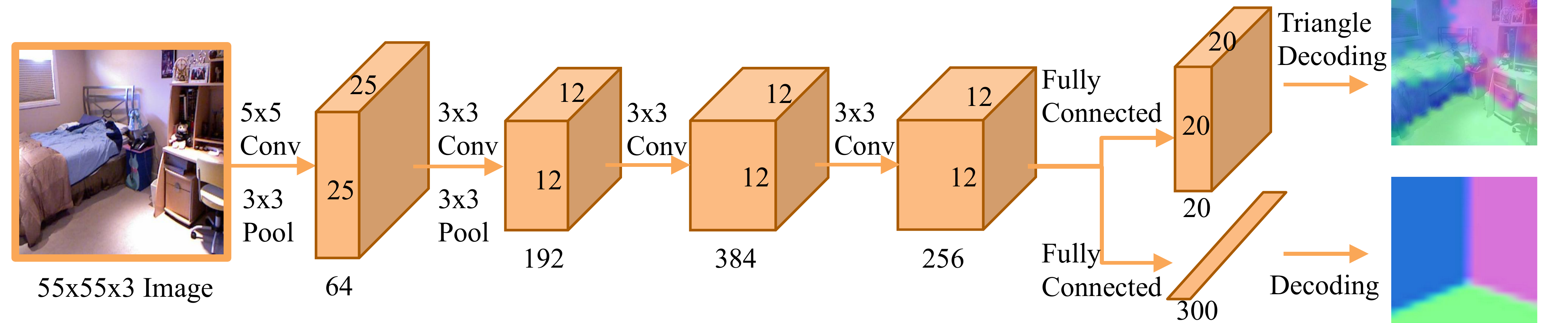}
\vspace{-0.2in}
\caption{\label{fig:coarse}The architecture of our top-down global network. Given an $55 \times 55$ image as input, it is passed though 4 convolutional layers. On the top of the last convolutional layer, the neurons are fully connected to two separate outputs: (i) global scene surface normals and (ii) room layouts.}
\vspace{-0.2in}
\end{figure*}

\vspace{-0.05in}
\subsection{Top-down Global Network}
\vspace{-0.05in}
\label{sec:topdown}
The goal of this network is to capture the coarse structure, enabling the interpretation of ambiguous portions of the image which cannot be decoded by local evidence alone.

\noindent {\bf Input:} Whole image rescaled to  $55 \times 55 \times 3$.

\noindent {\bf Output:}  Given the whole image as an input, we produce two complementary global interpretations as outputs: (i) a structural estimation of surface normals for the image and (ii) a cuboidal approximation of the image as introduced by \cite{Hedau09} and used in \cite{Lee10,Schwing12}, among others. For surface normal estimation, the output layer is $M_t \times M_t \times K_t $ where $M_t \times M_t$ is the size of output image for surface normals and $K_t$ is the number of classes uses in codebook.  For room layout, we use simple classification over 300 categories. We use $M_t=20, K_t = 20$.

\noindent {\bf Architecture:} The top-down global network includes four
convolutional layers; these layers are shared by the two tasks (surface normal and room layout estimation). The output of the neurons in the fourth convolutional layers are then fully connected to these two types of labels. To simplify the
description, we denote the convolutional layer as $C(k, s)$, which indicates the there are $k$ kernels, each having the size of $s \times s$. During convolution, we set all the strides to $1$.  We also denote the local response normalization layer as $LRN$, and the max-pooling layer as $MP$. The stride for pooling is $2$ and we set the pooling operator size as $3 \times 3$. Then the network architecture for the convolutional layers can be described as: $C(64,5 )  \rightarrow MP \rightarrow LRN \rightarrow C(192, 3 ) \rightarrow MP \rightarrow LRN \rightarrow C(384, 3) \rightarrow C(256, 3)$. For the surface normal estimation, neurons in the fourth convolutional layers are fully connected to the output space of $M_t \times M_t \times K_t$, which is $20 \times 20 \times 20 = 8000$. For the room layout estimation, we connect the same set of neurons to the $K_l=300$ labels. The architecture of the network is shown in Figure~\ref{fig:coarse}.

\noindent {\bf Loss function:} We treat both tasks as classification problems. For the room layout classification, we simply employ the softmax regression to define the loss.

For the surface normals estimation, we denote $F_i(I)$ as a $K_t$-class classification output for $i$th pixel on surface normal output map. We also apply softmax regression to optimize the function $F_i(I)$. Then the loss for the structural outputs of surface normals can be represented as,
\begin{eqnarray}\label{eq:multilr}
L(I,Y) &=& - \sum^{M \times M}_{i = 1} \sum^{K}_{k = 1} ( \mathbbm{1}(y_i = k) \log F_{i,k}(I)  ) ,
\end{eqnarray}
where $F_{i,k}(I)$ represents the probability that $i$th pixel should have surface normal defined by $k$th codeword, $\mathbbm{1}(y_i = k)$ is the indicator function, $Y = \{ y_i \}$ are the groundtruth labels for the  surface normals, $M = M_t$ and $K = K_t$.

During training, we learn the networks with these two losses simultaneously. As we have structural outputs for surface normals and only one prediction for the room layout, we need to balance the learning rate for both losses. Specifically, we have the learning rate $\sigma$ for optimizing the surface normals estimation, and the learning rate for layout estimation is set as $50 \sigma$.


\vspace{-0.05in}
\subsection{Bottom-up Local Network}
\vspace{-0.05in}
\label{sec:bottomup}

The goal of this network is to capture local evidence at a higher resolution that might be missed by the top-down network. We take a sliding window approach where we extract features in a window and predict the image properties in the center of the window. This sort of model has been applied successfully for generating local image interpretations in the form of normals \cite{Fouhey13a,Ladicky14b}
and semantic edges \cite{Dollar13}.

\begin{figure*}
\includegraphics[width=\textwidth]{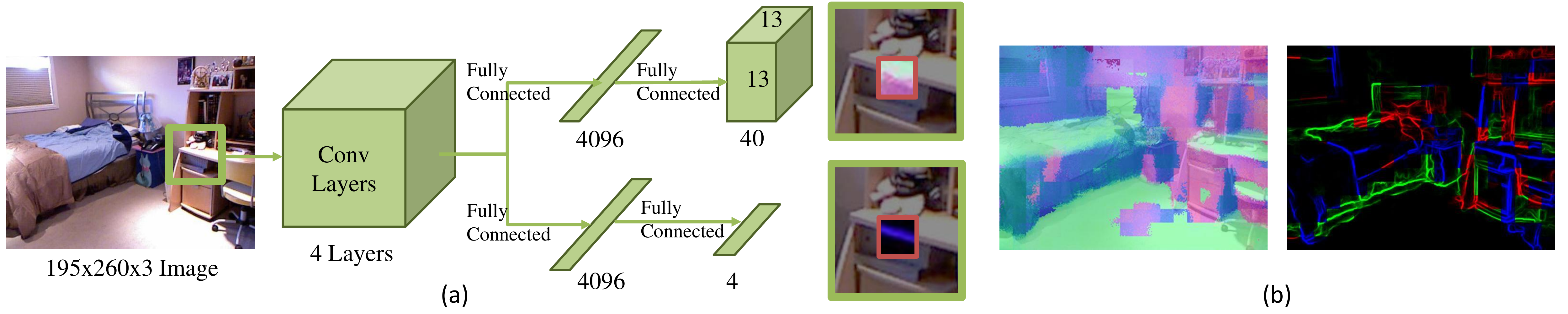}
\vspace{-0.3in}
\caption{\label{fig:bottom} Our bottom-up local network. (a) Given a full
image, we perform sliding window on it and extract $55 \times 55$ image
content as input. By forward propagation, the network produces the
local surface normals and convex/concave/occlusion edge labels for the middle $13 \times
13$ image patch; (b) After sliding window, we obtain the surface
normals and edge labels for the whole image. We plot the edge
labels on the output of Structured Edges \cite{Dollar13}. The colors blue, green and red represent
the convex, concave and occlusion edge labels, respectively.}
\vspace{-0.2in}
\end{figure*}

\noindent {\bf Input:} Given an image with size $195 \times 260$, we perform sliding window on it: the window size being $55 \times 55$ and stride of $13 $.

\noindent{\bf Output:} The local network produces two types of outputs: (i)
surface normals and (ii) an edge label. Each local sliding window predicts
the surface normal for $M_b \times M_b$ pixels at the center of the window.
We use $M_b=13$. As Figure~\ref{fig:bottom} illustrates,
our network takes a smaller part of the image as input and predicts the
surface normals in the middle of the patch, thus predicting the local normals
from local texture and its context.
We use $K_b=40$
codewords to define the output space. We use a larger number of
codewords since we expect local network to capture finer details. For the
edges, we use the classic categories of convex, concave, occlusion or not an edge. Note we just predict
one edge label for $13 \times 13$ pixels. For visualizations, we project
these edge labels onto Structured Edge~\cite{Dollar13}.

\noindent {\bf Architecture:}  The architecture of bottom-up network includes 4 convolutional layers and 2 sets of full connection layers. The convolutional layers are shared by the two tasks, and we use the same parameter settings mentioned in the top-down network. At the end of convolutional layers we stack two separate full connection layers with $4096$ neurons on it, each of which corresponds to one pipeline of task. For local surface normal estimation, we have the output size as $M_b \times M_b \times K_b = 13 \times 13 \times 40 = 6760$. On the other hand, there are $4$ outputs representing edge labels.

\noindent {\bf Loss Function:} Both of the tasks are defined as classification.
For edge label estimation, we apply softmax regression to define
the loss. For the local surface normal estimation we apply the loss defined in
Eq.\ref{eq:multilr} by setting $M = M_b$ and $K = K_b$. Similar to the coarse
network, we optimize these two tasks jointly during training, and the learning
rate for local surface normals and edge label estimation are $\sigma$  and $50
\sigma$, respectively.

\vspace{-0.05in}
\subsection{Visualization}
\begin{figure*}
\includegraphics[width=\textwidth]{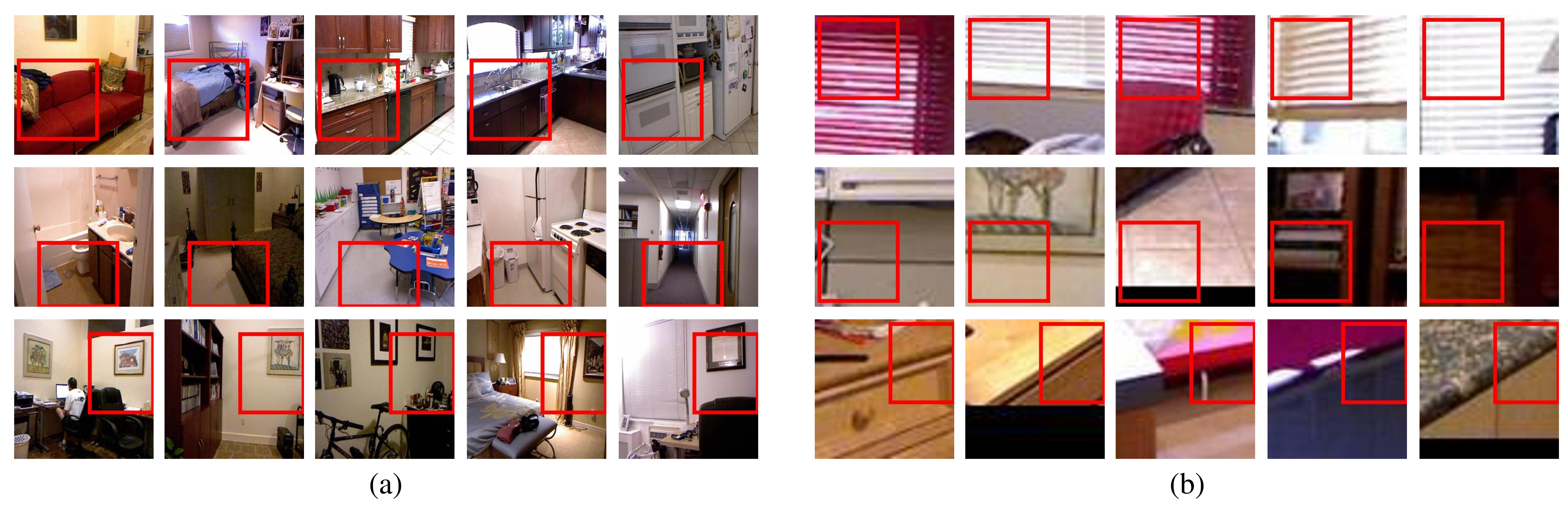}
\vspace{-0.3in}
\caption{\label{fig:fusion}Top regions for the 4th convolutional layer units in top-down and bottom-up networks. The receptive field for the neurons in the 4th layer is $31 \times 31$. We use red bounding boxes to represent the regions with top responses for different units. (a) The neurons from the top-down coarse network tempt to capture the structure information in the global scene; (b) The neurons from the bottom-up network response to the local texture and edges.}
\vspace{-0.15in}
\end{figure*}

\vspace{-0.05in}
We now attempt to analyze what the top-down and bottom-up networks learn.
Figure~\ref{fig:fusion} shows the top 5 activations for the units in the fourth
convolutional layer of (a) top-down network and (b) bottom-up network. Note
that these two networks share the same structure in convolutional layers, and
the size of receptive fields of the unit is $31 \times 31$. We select
representative samples for illustration. For the top-down global network, the
units capture high-level structures such as the side of beds, hallways and
paintings on the wall. For the bottom-up network, the units respond to local
texture and edges.

\vspace{-0.05in}
\subsection{Fusion Network}
\vspace{-0.05in}
\label{sec:fusion}

The goal of this network is to fuse the results of the two earlier networks and
refine their results. Each approach has complementary failure modes, and by
fusing the two networks, we show that better results can be obtained than
either by themselves. Additionally, both coarse and local network  treat every
pixel independently; our fusion network can also be thought of applying a form
of learned reasoning akin to \cite{Munoz10,Tu10} on our outputs.

\noindent {\bf Input:} As input to this fusion network, we take the output of
top-down and bottom-up networks; concatenate their outputs with input image and
use the concatenated features as input. The concatenation process
is as follows:

\begin{itemize}[noitemsep]
\item Global Coarse Output: The output of top-down coarse network is $20 \times 20$ with $20$ classes. We decode the output to a 3-dimensional continuous surface normal map and upscale it to $195 \times 260 \times 3$.
\item Layout: We select the room layout corresponding to the label with highest probability. The layout is a 3-channel feature map representing the surface normals in the layout. We resize it to $195 \times 260 \times 3$.
\item Local Surface Normals: The output of bottom-up local network in the sliding window format is $195 \times 260 \times 3$.
\item Edge Labels:  We obtain the 4 probabilities of edge labels per window. As the probabilities sum to $1$, we do not pass the no-edge output to the fusion network.
We upsample this three dimension vector to size $13 \times 13 \times 3$ for each window and obtain $195 \times 260 \times 3$ inputs.
\item Vanishing Point-Aligned Coarse Output: We adjust our coarse output's interpretation to match vanishing points estimated by \cite{Hedau09}, yielding another feature representation with the same size.
\end{itemize}
In addition to 15 channels described above, we also concatenate the original image and therefore our final input to the deep network is $195\times260\times18$.

\noindent {\bf Output:} The mid-level fusion network is also applied in a
sliding window scheme on the $195 \times 260$ image. By taking the inputs with
size of $55 \times 55$, we estimate the surface normals of the $M_b \times M_b$
center patch via the fusion network. Note that we use the same output window
size $M_b = 13$ as in the bottom-up local network, and the output space is defined
by $K_b = 40$ codewords.

\noindent{\bf Architecture:} The architecture of this network is stacked by 4 convolutional layers and 2 full connection layers. The convolutional layers also share the same parameter settings as the top-down and bottom-up networks. The last convolutional layer are fully connected to $4096$ neurons. These neurons further lead to the $13 \times 13 \times 40$ outputs which represent the surface normals. In testing time, we apply the fusion network on the feature maps with the stride of $M_b$.

\noindent {\bf Loss Function:} In training time, we fix the parameters of the top-down and bottom-up networks and obtain the feature maps of the training data though them. The loss function is defined as Eq.\ref{eq:multilr} by setting $M = M_b$ and $K = K_b$. To train the network, we apply the stochastic gradient descent with learning rate $\sigma$.

\vspace{-0.05in}
\section{Experiments}
\vspace{-0.05in}
\label{sec:experiments}
\vspace{-0.05in}
We now describe our experiments. We adopt the protocols used by state-of-the-art methods on this task \cite{Fouhey13a,Fouhey14c,Ladicky14b}.

\begin{figure*}
\includegraphics[width=\textwidth]{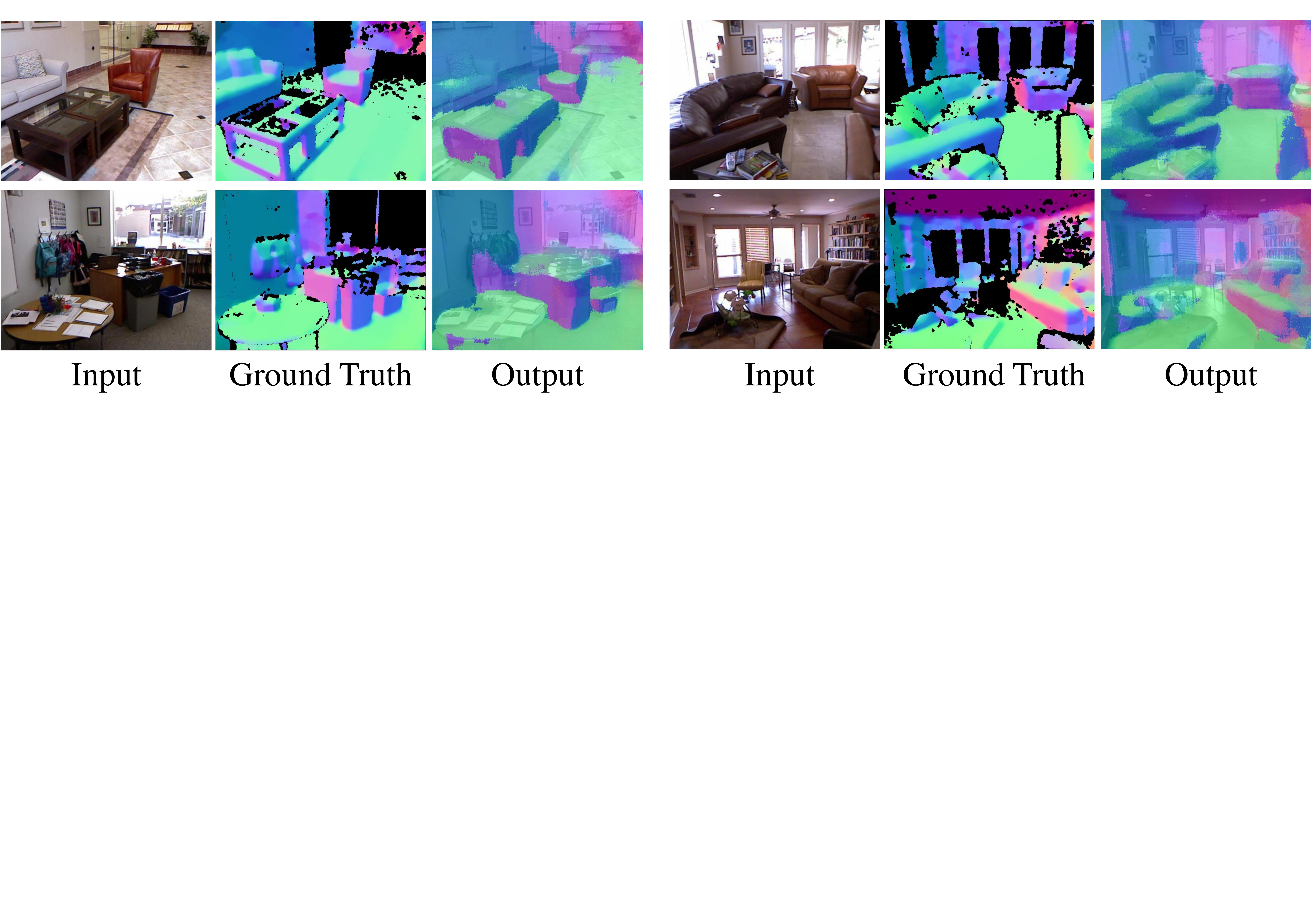}
\vspace{-0.3in}
\caption{\label{fig:gt1}Qualitative results of surface normal estimation using our complete architecture. Input images are shown on the left, ground truth surface normals from Kinect are shown in middle and the predicted surface normals are shown on right. Our network not only capture the coarse layout of the room but also preserves the fine details. Notice the fine details like the top of tables and couches, the legs of table are captured by our algorithm.}
\vspace{-0.1in}
\end{figure*}

\begin{figure*}
\includegraphics[width=\textwidth]{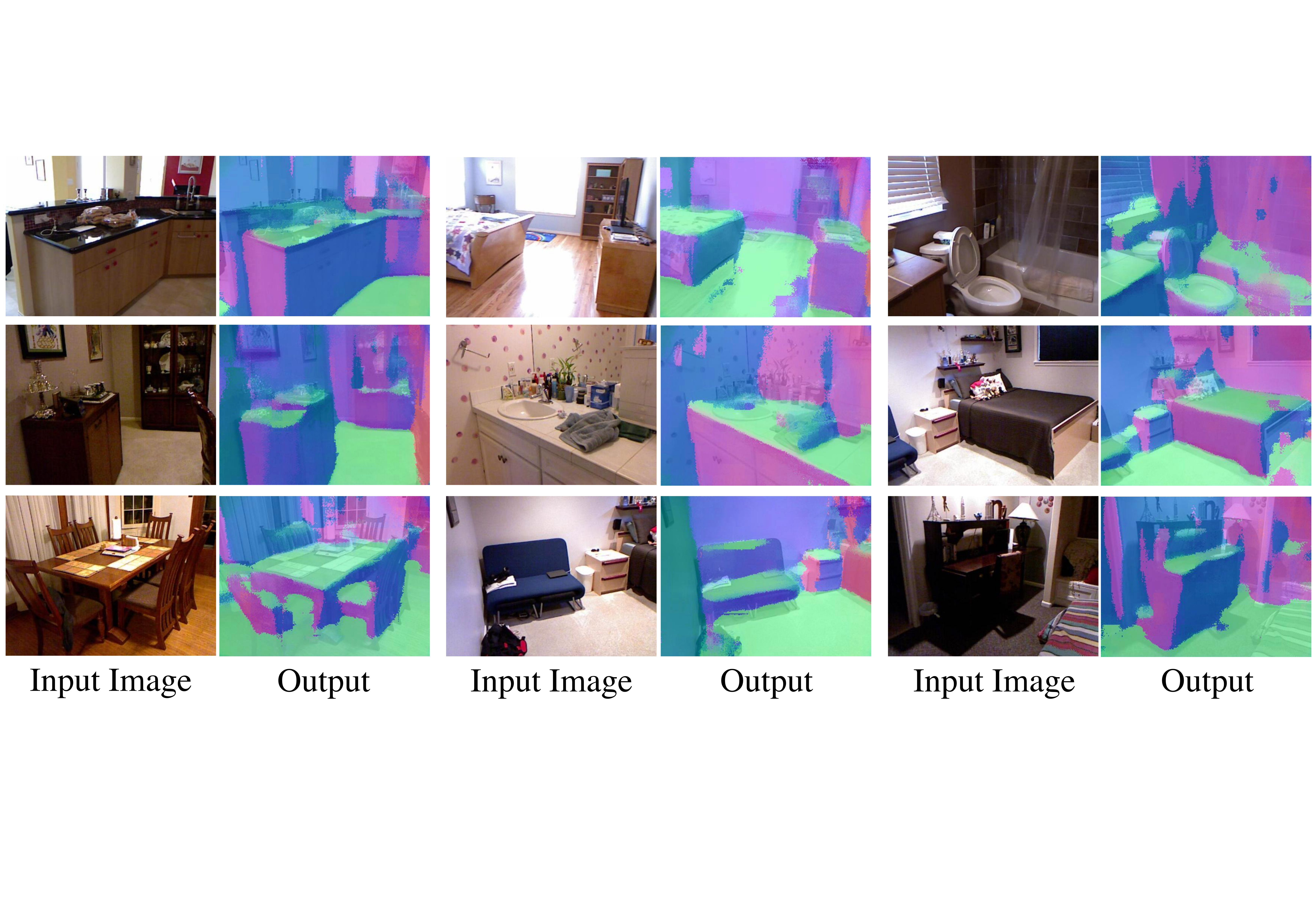}
\vspace{-0.2in}
\caption{\label{fig:gt2}More qualitative results to show the performance of our algorithm. Again notice the details captured such as the top of night-stands, the counters and even the legs of the chair are captured by our algorithm.}
\vspace{-0.15in}
\end{figure*}

\begin{figure*}
\includegraphics[width=\textwidth]{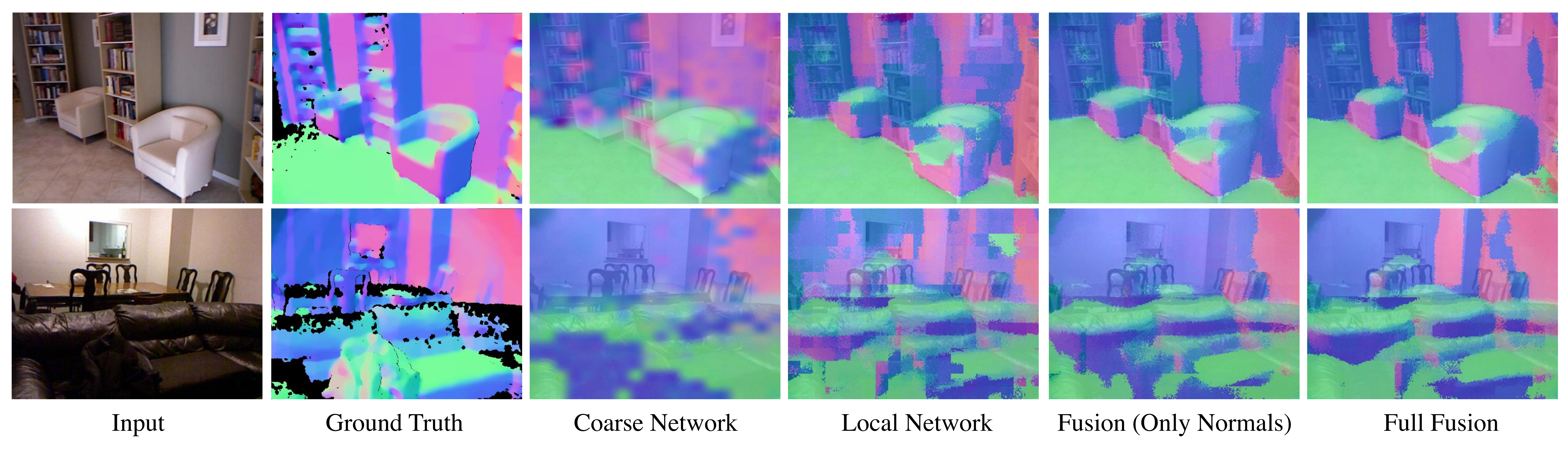}
\vspace{-0.25in}
\caption{\label{fig:ablative}Qualitative Ablative Analysis: The the global and local network estimation results have complementary failure modes. By combining both normal outputs we obtain better results via fusion network. With more information feeding in, the full fusion network reasons among them and improve the performance. }
\vspace{-0.15in}
\end{figure*}

\begin{figure*}
\centering
\includegraphics[width=\textwidth]{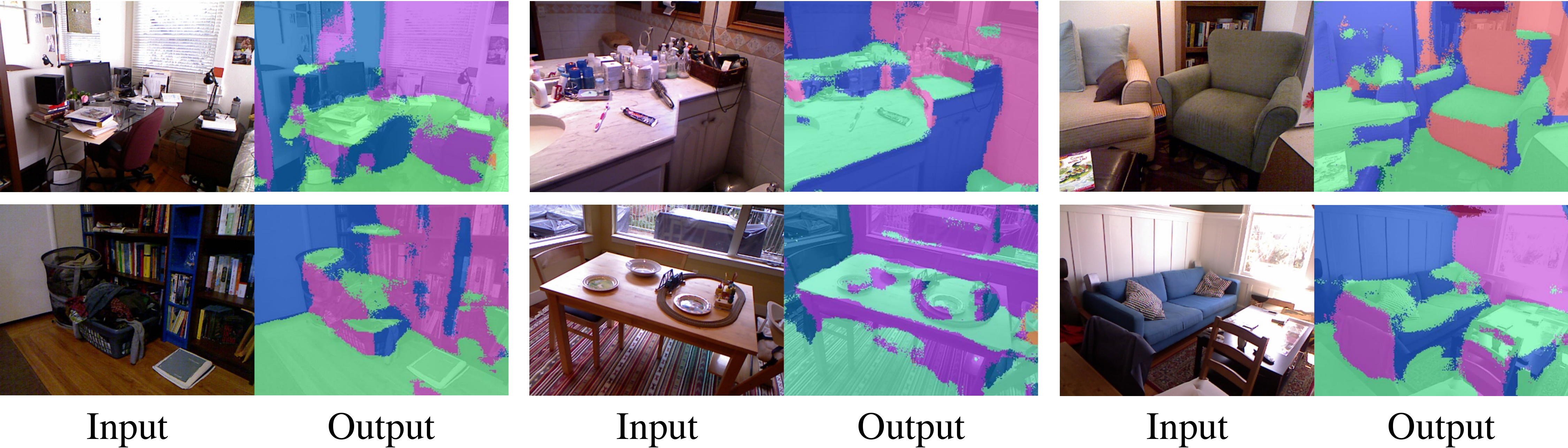}
\vspace{-0.25in}
\caption{Results on B3DO dataset\cite{Janoch11}. We obtain state of the art performance by applying our model trained on the NYU dataset without further fine-tuning on the B3DO dataset. }
\vspace{-0.2in}
\label{fig:gt3}
\end{figure*}

\noindent {\bf Dataset and Settings:} We evaluate our method on the NYU Depth
v2 dataset~\cite{Silberman12}. However, to train our models we use the
corresponding raw video data for the training images. We process the video data
using the provided development kit, but improve the normals with TV-denoising
as in \cite{Ladicky14b}. We apply the official split with $249$ scenes for
training and $215$ scenes for testing. In more details, we extract 200K frames
from the $249$ scenes for training, and test on the 654 images from the
standard test set. We also extract the room layout by fitting an inside-out box
to the the estimated surface normals~\cite{Hedau10}. The edge labels are
estimated using the ground-truth depth data in a procedure like \cite{Gupta13}.

During training, we fine-tune the network with stochastic gradient descent with
learning rate $\sigma = 1.0 \times 10^{-6}$. Note that during joint tuning with
the layouts and edges we set the learning rate as $50\sigma$ for these losses.
For training our coarse networks, we augment our data by flipping, color
changes and random crops. For training the local and fusion network, we rescale
the training images to $195 \times 260$ and randomly sample 400K patches with
size $55 \times 55$ from them.

\noindent {\bf Evaluation Criteria:} We adopt the criteria introduced in
\cite{Fouhey13a} to evaluate the surface normal prediction. We evaluate a
per-pixel error over the whole dataset, ignoring values that are unknown due to
missing depth data, as these often contain incorrect values. We summarize this population of per-pixel errors with
statistics: the mean, median, and RMSE, as well as percent-good-pixel metrics,
or what fraction of the pixels are correct within some threshold $t$ (for
$t = 11.25, 22.5, 30$).

\noindent {\bf Baselines:} Our primary baselines are the state-of-the-art in
surface normal prediction~\cite{Fouhey13a,Fouhey14c,Ladicky14b}. Each of these
is state-of-the-art in at least one metric that we evaluate on.

There have been no published results for the CNN performance on surface normal
layouts. Eigen et al.~\cite{Eigen14} recently proposed a two-level CNNs for
the task of depth prediction. In terms of quantitative results, they show that
the coarse CNN performs slightly worse that the two network architecture.
Therefore, as another baseline, we implemented  the regression coarse network of
\cite{Eigen14} with the loss changed to the negative dot-product.
Note that this network is a fully feedforward network with no intermediate
representations or designed structure. Therefore, it acts as a
good baseline for evaluation.

\vspace{-0.05in}
\subsection{Experimental Results}
\vspace{-0.05in}
\noindent {\bf Qualitative:} First, we demonstrate our qualitative results.
Figures~\ref{fig:gt1} and \ref{fig:gt2} show the results of our complete
architecture. Notice how our results capture the fine details of the input
image. Unlike many past approaches, our algorithm is able to correctly
estimate the surface normal of top of tables and chairs; it is able to estimate
the legs of the tables etc. Our algorithm is able to even estimate the top of
couch backs and how the surface normal changes across the couch (last
column, figure~\ref{fig:gt1}).

\noindent {\bf Quantitative:} Table~\ref{tab:eval} compares the performance of
our algorithm against several baselines. As the results indicate, our approach
is significantly better than all the baselines in all metrics. For many cases,
our results show as much as $15\%$ improvement over previously state of the art
   results.

\begin{table}
\centering
\caption{Results on NYU v2 for per-pixel surface normal estimation, evaluated over valid pixels. }
\vspace{-0.1in}
\label{tab:eval}
\begin{tabular}{@{}l@{ }c@{ }c@{ }c@{ }c@{ }c@{ }c@{ }c} \toprule
        & \multicolumn{3}{c}{Summary Stats. (${}^\circ$)}& \multicolumn{3}{c}{\% Good Pixels} \\
        & \multicolumn{3}{c}{(Lower Better)} & \multicolumn{3}{c}{(Higher Better)} \\
        & Mean & Median & RMSE & $11.25^{\circ}$ & $22.5^{\circ}$ & $30^{\circ}$ \\ \midrule
Our Network           & \bf 25.0     & \bf 13.8      & \bf 35.9      & \bf 44.2      & \bf 63.2  & \bf 70.3
\\
UNFOLD \cite{Fouhey14c} & 35.1      & 19.2      & 48.7      & 37.6      & 53.3  & 58.9 \\
Discr. \cite{Ladicky14b}    & 32.5      & 22.4      & 43.3      & 27.4  & 50.2  & 60.2 \\
3DP (MW) \cite{Fouhey13a} & 36.0      & 20.5      & 49.4      & 35.9      & 52.0  & 57.8 \\
3DP \cite{Fouhey13a}        & 34.2      & 30.0      & 41.4      & 18.6   & 38.6  & 49.9
\\ \bottomrule
\end{tabular}
\end{table}

\noindent {\bf Ablative Analysis:} Next, we perform a comprehensive ablative analysis to explore which component
of the network helps in improving performance. This ablative analysis should
also help us to verify the hypothesis that designing networks based on
meaningful intermediate representations and constraints can help improve the
performance.

First, we discuss some qualitative results shown in Figure~\ref{fig:ablative}.
As seen in the figure, the top-down coarse network just captures the coarse
structure of the room. For example, in the top figure, it misses the vertical
surface on the inner side of the couch or it misses how the vertical
orientations change due to bookshelves between couches. On the other hand, a
local network indeed captures those details. However, since it only observes local
patches, it completely misclassifies the wall patches below the picture frame.
Fusing the two networks preserves the finer details (inner side of
the couch and changing vertical orientations of the wall), but still
misclassifies a big patch on the wall near the picture frame. However, once
the network uses the edge labels (e.g., the convex edge of
the shelf and the missing edge on the wall) to improve the boundaries.

Quantitatively, we compare all the components one by one in
Table~\ref{tab:eval2}. The fusion network which combines the raw images,
surface normal predictions from bottom-up and top-down network provides a
significant boost in performance. Furthermore, adding layout (+Layout),
edges (+Edge) and vanishing points (+VP) independently improve the
performance of the network. By combining all of them together in the full
fusion network, we obtain better results (especially in the median error
and $11.25^\circ$ error).

As noted by the authors of \cite{Ladicky14b}, the triangular decoding scheme does not optimize
for mean error and RMSE well; we thus also report results with a
scheme that better optimizes these metrics (Soft): we weight the $40$ codewords with the output
distribution and calculate the weighted mean vector. Note
this does not require any retraining.

Finally, we note that our performance is significantly better than the
coarse network of Eigen et al.~\cite{Eigen14} implemented by us. And,
even the \cite{Eigen14} network performance is improved as we combine
the network and our bottom-up local network with the fusion network.


\begin{table}
\centering
\caption{Ablative Analysis}
\vspace{-0.1in}
\label{tab:eval2}
\begin{tabular}{@{}l@{ }c@{ }c@{ }c@{ }c@{ }c@{ }c@{ }c} \toprule
        & Mean & Median & RMSE & $11.25^{\circ}$ & $22.5^{\circ}$ & $30^{\circ}$ \\ \midrule
Full           &  25.0     & \bf 13.8      &  35.9      & \bf 44.2      & \bf 63.2  & \bf 70.3
\\
Full (Soft)           & \bf 24.2     &  17.3      & \bf 32.2      &  36.8      &  58.5  &  68.7
\\
Fusion (+VP)   &   25.3      &  14.4      &  35.9     &  42.7      &  62.5 &  69.9
\\
Fusion (+Edge)   &   25.8      &  15.3      &  36.0     &  40.0      &  61.6 &  69.7
\\
Fusion (+Layout)   &   25.8      &  14.9      &  36.3     &  41.1      &  61.9 &  69.5
\\
Fusion      &   26.0      &  15.5      &  36.2     &  39.5      &  61.3 &  69.3 \\
Bottom-up     &  32.2      &  23.5      &  42.0      &  27.2     &  48.5  &   58.5 \\
Top-down        &  29.0      &  19.8      &  38.3      &  32.7      &  53.8  &  62.4 \\
\\
Eigen et al.(Fusion)       &  26.8     &  19.3      &   35.2      &   32.6      &  55.3  &  65.5 \\
Eigen et al.(Coarse)      &  27.9     &  23.4      &  34.5      &  25.5      &  48.4  &   60.6 \\
\bottomrule
\end{tabular}
\end{table}

\vspace{-0.1in}
\subsection{Berkeley B3DO Dataset}
\vspace{-0.05in}
To show our model can generalize well, we apply it directly
on the B3DO \cite{Janoch11} dataset. We note
that there is significant mismatch in dataset capture between
the two: NYU contains almost exclusively full scenes while
the B3DO contains many close-up views. Since B3DO also contains
many scenes with downwards facing views, we rectify our
results to detected vanishing points to compensate.
We report our
results of our full fusion network in Table~\ref{tab:eval3}, which shows that
our method outperforms the baselines from \cite{Fouhey13a} by a nice margin
in all metrics. We show some qualitative results in Figure~\ref{fig:gt3}.

\begin{table}
\centering
\caption{B3DO}
\vspace{-0.1in}
\label{tab:eval3}
\begin{tabular}{@{}l@{ }c@{ }c@{ }c@{ }c@{ }c@{ }c@{ }c} \toprule
        & Mean & Median & RMSE & $11.25^{\circ}$ & $22.5^{\circ}$ & $30^{\circ}$ \\ \midrule
Full           & \bf 34.5      & \bf 20.1      & \bf 47.9      & \bf 36.7      & \bf 52.4  & \bf 59.2
\\
3DP(MW)~\cite{Fouhey13a}         & 38.0     & 24.5      &  51.2      &  33.6      &  48.5  &  54.5
\\
Hedau et al.~\cite{Hedau09}   &   43.5      &  30.0      &  58.1     &  32.8      &  45.0 &  50.0
\\
Lee et al.~\cite{Lee09}   &   41.9      &  28.4      &  56.6     &  32.7      &  45.7 &  50.8
\\
\bottomrule
\end{tabular}
\end{table}

\vspace{-0.05in}
\section{Conclusion}
\vspace{-0.05in}
We have presented a novel CNN architecture for surface normal estimation. By injecting the 3D insights learned in the past ten years, our model reaches the state of the art performance. Qualitatively, our model works surprisingly well and  can even capture fine details such as table legs, curved surfaces of couches etc.

\begin{footnotesize}
\textbf{Acknowledgments:} This work was partially supported by NSF IIS-1320083, ONR MURI N000141010934, Bosch Young Faculty Fellowship to AG and NDSEG fellowship to DF. This material is also based on research partially sponsored by DARPA under agreement number FA8750-14- 2-0244. The U.S. Government is authorized to reproduce and distribute reprints for Governmental purposes notwithstanding any copyright notation thereon. The views and conclusions contained herein are those of the authors and should not be interpreted as necessarily representing the official policies or endorsements, either expressed or implied, of DARPA or the U.S. Government. The authors wish to thank NVIDIA for the donations of GPUs.
\end{footnotesize}


{\small
\bibliographystyle{ieee}
\bibliography{local}
}

\end{document}